\documentclass[letterpaper]{article} 
\usepackage{aaai25}  
\usepackage{times}  
\usepackage{helvet}  
\usepackage{courier}  
\usepackage[hyphens]{url}  
\usepackage{graphicx} 
\usepackage{natbib}  
\usepackage{caption} 
\usepackage{algorithm}
\usepackage{algorithmic}

\usepackage{newfloat}
\usepackage{listings}
\usepackage{amsmath}
\usepackage{booktabs}
\usepackage{multirow}

\usepackage{amssymb}
\DeclareCaptionStyle{ruled}{labelfont=normalfont,labelsep=colon,strut=off} 
\floatstyle{ruled}
\newfloat{listing}{tb}{lst}{}
\floatname{listing}{Listing}
\title{External Reliable Information-enhanced Multimodal Contrastive Learning \\ for Fake News Detection}

\title{External Reliable Information-enhanced Multimodal Contrastive Learning \\ for Fake News Detection}

\author{
    Biwei Cao\textsuperscript{\rm 1, \rm 2}, 
    Qihang Wu\textsuperscript{\rm 1, \rm 2}, 
    Jiuxin Cao\textsuperscript{\rm 1, \rm 2, \rm 3}\thanks{Corresponding authors},
    Bo Liu\textsuperscript{\rm 2, \rm 3, \rm 4}, 
    Jie Gui\textsuperscript{\rm 1, \rm 2, \rm 3, \rm 5}\footnotemark[1]
}
\affiliations {
    \textsuperscript{\rm 1} School of Cyber Science and Engineering, Southeast University, Nanjing 211189, China\\
    \textsuperscript{\rm 2} Key Laboratory of Computer Network and Information of Ministry of Education of China, Nanjing 211189, China\\
    \textsuperscript{\rm 3} Purple Mountain Laboratories, Nanjing 210000, China\\
    \textsuperscript{\rm 4} School of Computer Science and Engineering, Southeast University, Nanjing 211189, China\\
    \textsuperscript{\rm 5} Engineering Research Center of Blockchain Application, Supervision And Management (Southeast University), Ministry of Education\\
    
    \{caobiwei, wuqihang, jx.cao, bliu, guijie\}@seu.edu.cn 
}

\usepackage{bibentry}

\begin{document}

\maketitle

\begin{abstract}
With the rapid development of the Internet, the information dissemination paradigm has changed and the efficiency has been improved greatly. While this also brings the quick spread of fake news and leads to negative impacts on cyberspace. Currently, the information presentation formats have evolved gradually, with the news formats shifting from texts to multimodal contents. As a result, detecting multimodal fake news has become one of the research hotspots. However, multimodal fake news detection research field still faces two main challenges: the inability to fully and effectively utilize multimodal information for detection, and the low credibility or static nature of the introduced external information, which limits dynamic updates. To bridge the gaps, we propose ERIC-FND, an external reliable information-enhanced multimodal contrastive learning framework for fake news detection. ERIC-FND strengthens the representation of news contents by entity-enriched external information enhancement method. It also enriches the multimodal news information via multimodal semantic interaction method where the multimodal constrative learning is employed to make different modality representations learn from each other. Moreover, an adaptive fusion method is taken to integrate the news representations from different dimensions for the eventual classification. Experiments are done on two commonly used datasets in different languages, X (Twitter) and Weibo. Experiment results demonstrate that our proposed model ERIC-FND outperforms existing state-of-the-art fake news detection methods under the same settings. 

\end{abstract}


\section{Introduction}

With the rapid development and widespread use of the internet, the number of internet users has been increasing, making reading news online become an important part of daily activities. The Internet provides great convenience for disseminating information on social network platforms, leading to a large amount of fake news published on the Internet for various intentions. The widespread dissemination of fake news does not only severely destroy the credibility and social influence of mainstream media but also has the potential to cause panic in society. To mitigate the negative impact caused by fake news, automated fake news detection has gradually become a research hotspot.

Early fake news detection mainly focused on propagation structures or text-based detection\cite{rw:2t}. With the development of social network platforms, the news format has gradually shifted from pure text to multimodal content. Correspondingly, the research on fake news detection has changed to multimodal fake news detection as well. Existing multimodal fake news detection methods mainly focus on how to integrate the news representation from multi-modalities to improve detection performance, instead of doing in-depth research on the inconsistencies between images and texts in fake news. Furthermore, the most multimodal methods rarely explore external information and the external information used often lacks reliability or remains static \cite{rw-c-m-4}.

To address these issues, we propose a method that aligns news text and image information to a shared feature space through multimodal contrastive learning. Then, it obtains enhanced multimodal representations via cross-modal semantic interaction. Additionally, to deepen the model understanding of news text, descriptions of knowledge entities from Wikipedia are introduced as external reliable information. An external information enhancement module is employed, utilizing an attention mechanism to enhance the feature representation of news content with the external reliable information. Our major contributions are summarized as follows.

\begin{itemize}
\item We propose ERIC-FND, an External Reliable Information-enhanced multimodal Contrastive learning method for Fake News Detection, which mainly consists of three modules, namely external information enhancement module, multimodal information interactive enhancement module and adaptive fusion-based classification module. The first two modules aim to leverage external information for a deeper understanding of news content and to enhance multimodal representation based on multimodal information. The adaptive fusion based classification module attempts to adaptively fuse the obtained features of different dimensions, improving the classification performance and interpretability.

\item To achieve a better representation of news incorporating both images and text, we utilize multimodal contrastive learning to align the news textual features with its visual features. We then use cross-modal semantic interaction method to capture the deep interactive semantic features. What is more, to enhance the model understanding of the news contents, entity-enriched information integration method is proposed to obtain the news-related external information and use this information to enhance the feature representation of news.

\item Extensive experiments are conducted on two datasets to demonstrate the performance of the proposed model. The results show that our model outperforms SOTA methods.
\end{itemize}

\section{Related Work}

In the early stage, the methods are mainly based on traditional machine learning, which focuses on feature engineering to extract static features from the post propagation processes \cite{rw1,rw2,rw3,rw4,rw5}. With the development of deep learning, fake news detection research shifts towards the methods based on deep learning, due to its strong ability of handling complex data and pattern recognition. Currently, the research directions for fake news detection based on deep learning can be generally divided into two types: social context-based fake news detection and content-based fake news detection \cite{rw:2t}.

\subsection{Propagation Structure-based Fake News Detection}
Fake news detection methods based on social context mostly utilize the characteristics and patterns of information transmission on social media platforms to detect fake news \cite{rw-p-1}. Researchers believe that fake news exhibits different propagation patterns and network structures compared to real news during the dissemination process \cite{rw-p-2}. Ma et al. \citeyearpar{rw-p-3} model the posts diffusion with propagation trees and propose a Propagation Tree Kernel method which calculates the similarity between the propagation tree structures of different rumors. Liu et al. \citeyearpar{rw-p-4} applies recurrent and convolutional networks to capture both global and local variations of user characteristics along propagation paths to detect fake news. Sun et al. \citeyearpar{rw-p-5} propose a novel hyperedge walking strategy on a meta-hyperedge graph for gaining news propagation sub-structure representations in social network. To relieve the problem of lacking in labelled dataset, Fang et al. \citeyearpar{rw-p-6} propose a tree VAE-based sentiment propagation module to leverage the propagation structure. Lin et al. \citeyearpar{rw-p-7} represent social media rumors as diverse propagation threads and design a hierarchical prompt encoding mechanism to learn contextual representations independent of language. The Rumor Adaptive Graph Contrastive Learning (RAGCL) \cite{rw-p-7} method is proposed to improve rumor detection by adapting graph-based learning of rumor propagation tree structures and focusing on key substructures via adaptive view augmentation.

\subsection{Content-based Fake News Detection}
\subsubsection{ Unimodal Fake News Detection}
The content-based fake news detection starts with the unimodality. Ma et al. \citeyearpar{rw-c-0} first apply deep learning to fake news detection. Since the good sequential data modeling capabilities of recurrent neural networks (RNNs), Ma et al. employ RNNs to learn hidden layer representations and capture the content changes of related texts over time. In 2017, Yu et al. \citeyearpar{rw-c-1}  are the first to model news articles using convolutional neural networks (CNNs) and propose the CNN-based CAMI method. This method maps tweets related to a news event into a vector space, concatenates them into a matrix, and then uses a CNN to extract textual features. Inspired by the adversarial learning from Generative Adversarial Networks (GAN), Ma et al. \citeyearpar{rw-c-2} design a generator to produce noise, making the original conversational threads more complex, thus forcing the discriminator to learn stronger rumor indicative representations from difficult samples. For more information for fake news detection, several works exploit external knowledge. Hu et al. \citeyearpar{rw-c-3} propose an end-to-end graph neural model called CompareNet, which compares the knowledge base (KB)-based entity representations with the news contextual entity representations to capture consistency between the news content and the KB.  With Combination of the user comments and user information, Tseng et al. \citeyearpar{rw-c-4} establish fact-based associations with entities in the news content. What is more,  researchers also focus on the fake image detection for the news \cite{rw-c-i-1,rw-c-i-2,rw-c-i-3}. 
\begin{figure*}[t]
\centering
\includegraphics[width=0.8\textwidth]{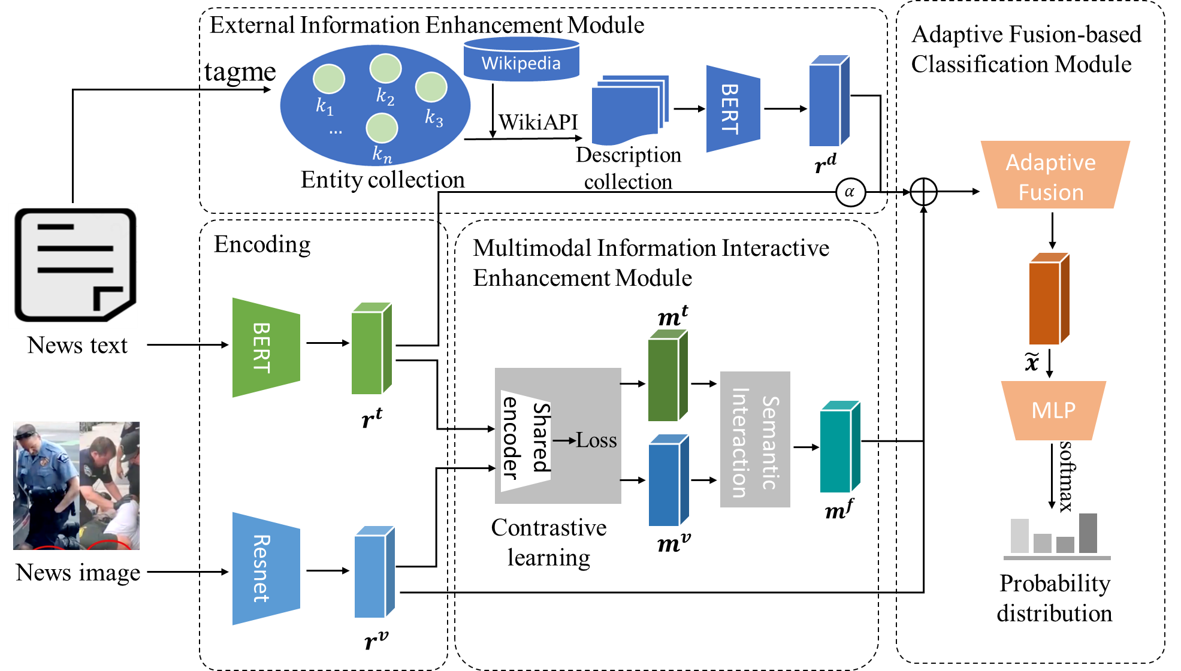} 
\caption{The proposed model ERIC-FND framework, where 
\textcircled{a} represents the attention mechanism and \textcircled{+} represents the concatenation operation.}
\label{framework}
\end{figure*}

\subsubsection{ Multimodal Fake News Detection}
With the development of information technology, multimodal news formats have become mainstream. Jin et al. \cite{rw-c-m-1} first use deep learning network for multimodal fake news detection. The att-RNN is proposed to combine the multimodal features including images, texts and social context.  Khattar et al. \cite{rw-c-m-2} use a bimodal variational autoencoder coupled with a binary classifier to learn probabilistic latent variable models. Chen et al. \cite{rw-c-m-3} propose an ambiguity-aware multimodal fake news detection method which estimate the multimodal ambiguity and capture the cross-modal correlations. Hu et al. \cite{rw-c-m-4} extract the rumor evidence of different modalities from the Internet and construct a multimodal dataset with the Internet search results. However, the extensive searching on the internet introduces a large amount of noise. Zhang et al. \cite{rw-c-m-5} utilize reinforcement learning to build a knowledge subgraph for each news to keep the useful knowledge related to news.

\section{Problem Formulation}
\textbf{Input}: Given the news data set $S$, and $n$ pieces of data, we have

$
S=\left\{\left(t_{1}, v_{1}, l_{1}\right), \ldots,\left(t_{k}, v_{k}, l_{k}\right), \ldots,\left(t_{n}, v_{n}, l_{n}\right)\right\}
$, where$\left(t_{k}, v_{k}, l_{k}\right) \in S$ is the $k$th piece data in the dataset, $t_{k}$ denotes the $k$-th piece of news text, $v_{k}$ denotes the image corresponding to the $k$-th piece of news text, $l_k \in \{0,1\}$ denotes the label of the $k$-th news item, and one piece of news corresponds to one label.

\textbf{Output}: A prediction label for the news, either fake news or true news.

\section{Methodology}
The framework of proposed ERIC-FND model is illustrated in Figure ~\ref{framework}. The model has three main modules: External Information Enhancement Module to enhance the news textual content based on the external information, Multimodal Information Interactive Enhancement Module to achieve the combined multimodal representation, and Adaptive Fusion-based Classification Module to adaptively fuse all the features to  complete the classification.

\subsection{Encoding}
We first do the feature encoding separately for the text and image of a piece of news.

For text part, we apply BERT model as the encoder on the pre-processed texts and input the vector of $[CLS]$ in the last hidden layer into the fully connected layer to obtain the representation of news text $r^t$, which can be formulated as follows:
\begin{equation}
    r^{cls} = Text\_Encoder(t),
\end{equation}
\begin{equation}
    r^t = W_{tf} \cdot r^{cls},
\end{equation}
where $r^t \in \mathbb{R}^{n \times l \times d_h} $, $l$ is the text length and $d_h$ is the dimensionality of hidden layer. $W_{tf}$ is the weight matrix of fully connected layer during the news text encoding process.

For image part, we start with resizing of image into the standard size. Then, the convolutional layers of ResNet-50 are used and a fully connected layer is added after the last convolutional layer. The output from the fully connected layer is taken as the representation of the news image $r^v$ which can be formulated as follows:

\begin{equation}
    r^{vis} = Visual\_Encoder(t),
\end{equation}
\begin{equation}
    r^v = W_{tf} \cdot r^{vis},
\end{equation}
where  $r^v \in \mathbb{R}^{n \times d_v} $ and $d_v$ is the dimensionality of image feature. $W_{vf}$ is the weight matrix of fully connected layer during the news image encoding process.

\subsection{External Information Enhancement Module}

External information enhancement module consists of two parts, namely external information extraction and information enhancement which takes entity-enriched external information enhancement method to improve the news text representation by integrating with external reliable information. 
\subsubsection{External Information Extraction}
In the beginning, we extract the entity from news textual content to explore the entities that have the potential to be knowledge. Since X and Weibo datasets which are in different languages are selected in implementation, for Chinese, jieba segmentation tool is employed to classify the parts of speech in the news and extract noun entities from the news text, while for English, tagme tool is selected to extract all the concept entities. The knowledge concept entity collection obtained can be formulated as $E = \{k_1,k_2,k_3, \ldots, k_i, \ldots, k_n\}$, where $k_i$ stands for the $i$th concept entity and $n$ represents the number of entities.
\begin{figure}[t]
\centering
\includegraphics[width=0.4\textwidth]{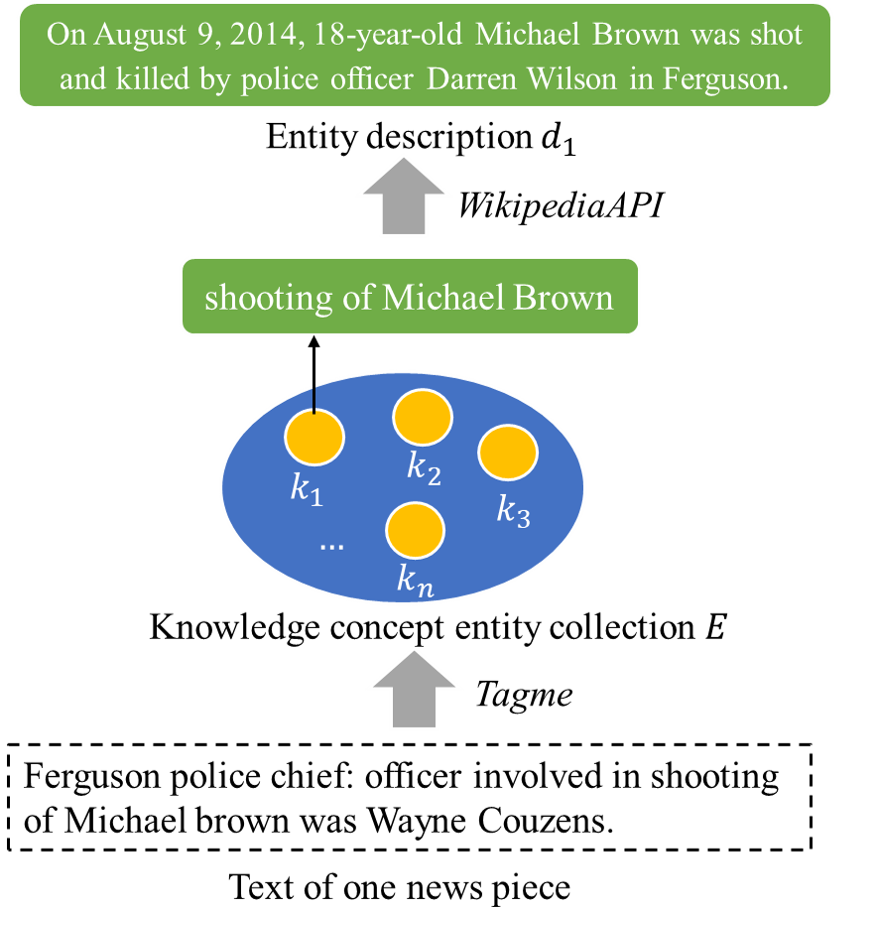} 
\caption{The process of entity description extraction.   }
\label{desc}
\end{figure}

Then, we link the news entities with Wikipedia. The extracted knowledge entities from the news text are mapped to Wikipedia and the corresponding entity descriptions are retrieved. The specific process for obtaining entity descriptions is shown in Figure ~\ref{desc}. The WikipediaAPI is taken to obtain the description of the knowledge concept entity “shooting of Michael Brown” from Wikipedia.

In practice, we observe that the entity descriptions returned by Wikipedia are usually long texts, where the first sentence often provides an essential conceptual description of the entity. Therefore, to avoid noise from the long entity descriptions, this study uses only the first sentence of the entity description. Additionally, when analyzing news texts, multiple entities are often identified. Thus multiple entity descriptions are retrieved from Wikipedia. Therefore, the collection $D$ of all entity descriptions in a piece of news can be represented as $D = \{d_1, d_2, \ldots, d_i, \ldots, d_n\}$, where $d_i$ is the description of the $i$th entity in Wikipedia, and $n$ denotes the number of entity descriptions.

\subsubsection{ Attention-based Information Enhancement }
In this part, we do the feature extraction on the entity description collection and integrate the context feature of external information with news content. 

Similar to news text encoding, we apply BERT to obtain the contextual representation of news entity description. For an entity description $d$, the process of getting the contextual feature $r^d$ can be formally represented as
\begin{equation}
    r^{desc} = Text\_{Encoder}(d),
\end{equation}

\begin{equation}
    r^{d} = W_{df} \cdot r^{desc},
\end{equation}
where $W_{df}$ represents the weight matrix of the fully connected layer. As a result, we can achieve the entity description feature matrix of news text $M^d$. 

Then, the representation of entity description is enhanced based on news text via attention mechanism. We calculate the dot product between news text feature $r^t$ and the entity description feature matrix $M^d$ as the attention weight $attF$. Next, use $attF$ to extract key features from the  entity description matrix, and obtain the key entity description feature $\tilde{r}^{desc} $ by a mean pooling operation. This calculation process can be formulated as follows:
\begin{equation}
    attF=softmax(r^t M^d),
\end{equation}
\begin{equation}
    \tilde r^{desc} = Mean\_Pooling(attF \odot M_d).
\end{equation}

Finally, to further enhance the representation ability of news text features, we combine the obtained entity description feature $\tilde{r}^{desc}$ with the  news text feature $r^t$ through an additive fusion operation. This process results in external information enhanced text features $\tilde{r}^t$: 

\begin{equation} \label{eq1}
    \tilde{r}^t = W_t r^t + W_d \tilde{r}^{desc}.
\end{equation}

\subsection{Multimodal Information Interactive Enhancement Module}
This module has two main parts: multimodal contrastive learning and cross-modal semantic interaction. We align the multimodal news informaion through multimodal contrastive learning and gain the enhanced news representation based on the different modal features via  cross-modal semantic interaction.

\subsubsection{Multimodal Contrastive Learning}

Feature alignment helps balance the contributions of different modalities and reduces the loss of information. Here we apply cross-modal contrastive learning to map each unimodal feature to a shared representation space to achieve alignment of different modal features.

In the training of multimodal contrastive learning, for the batch with $N$ examples $x = \{(x^v_i,x^t_i)\}^N_{i=1}$, there are $N \times N$ combinations of images and texts, including $N$ positive pairs and $N^2-N$ negative pairs, where $x^v_i$ and $x^t_i$ are the image and text of the $i$th pair. The training objective  of contrastive learning is to align visual feature representations and textual feature  representations by ensuring positive pairs are closer in the shared representation space compared to negative pairs.

Specifically, we first map the news text representation $r^t$ and visual representation $r^v$ to a shared vector space and use the L2 norm to get the text vectors $e^t$ and visual vector $e^v$ with the same dimensionality, which can be formulated as

\begin{equation}
    e^{t} = Shared\_Encoder(r^t),
\end{equation}

\begin{equation}
    e^{v} = Shared\_Encoder(r^v),
\end{equation}
where the shared encoders for text and image are fully connected layer structures to separately map the unimodal representations into shared vector space.

For the $i$th image and $j$th text in the batch, the similarity, which reflects the degree of semantic consistency between the image and text, is denoted as $p_{ij}^{v \rightarrow t}$. The calculation of the similarity is  shown as 
\begin{equation}
    p_{ij}^{v \rightarrow t} = \frac{\exp(sim(e_i^v, e_j^t) / \tau)}{\sum_{j=1}^N \exp(sim(e_i^v, e_j^t) / \tau)},
\end{equation}
where $sim(\cdot)$ represents the dot product operation. $\tau$ is a hyperparameter of the temperature coefficient. The larger the value of $\tau$ is, the less the model penalizes difficult examples.

Similarily, the calculation of the similarity between the $i$th text and $j$th image in batch is:
\begin{equation}
    p_{ij}^{t \rightarrow v} = \frac{\exp(sim(e_i^t, e_j^v) / \tau)}{\sum_{j=1}^N \exp(sim(e_i^t, e_j^v) / \tau)}.
\end{equation}
where $sim(\cdot)$ represents the dot product operation. $\tau$ is a hyperparameter of the temperature coefficient. 

We use the InfoNCE loss for training. The loss function $L^{v \rightarrow t}$ between images and texts can be formally represented as

\begin{equation}
    L^{v \rightarrow t} = -\frac{1}{N} \sum_{i=1}^N \sum_{j=1}^N y_{ij}^{v \rightarrow t} \log p_{ij}^{v \rightarrow t}.
\end{equation}

Correspondingly, we have the loss function $L^{t \rightarrow v}$ between texts and images:
\begin{equation}
    L^{t \rightarrow v} = -\frac{1}{N} \sum_{i=1}^N \sum_{j=1}^N y_{ij}^{t \rightarrow v} \log p_{ij}^{t \rightarrow v}.
\end{equation}

The overall contrastive loss function $L_c$ is the average of $L^{v \rightarrow t} $ and $L^{t \rightarrow v}$:

\begin{equation}
    L_c = \frac{1}{2} \left( L^{(v \rightarrow t)} + L^{(t \rightarrow v)} \right).
\end{equation}

\subsubsection{Cross-modal Semantic Interaction}
This part learns the deep semantic connections between different modalities. Given the aligned visual representation $m^v$ and aligned text representation $m^t$, we first calculate the attention weights between different modality representations $f_{v \rightarrow t}$ and $f_{t \rightarrow v}$ which can be formulated as

\begin{equation}
    f_{t \rightarrow v} = softmax \left(  \frac{\left[m^t\right]\left[ m^v \right]^T}{\sqrt{dim}}   \right),
\end{equation}

\begin{equation}
    f_{v \rightarrow t} = softmax \left(  \frac{\left[m^v\right]\left[ m^t \right]^T}{\sqrt{dim}}   \right),
\end{equation}
where $dim$ demotes the dimensionality.

Then, the attention weights  $f_{v \rightarrow t}$ and $f_{t \rightarrow v}$  are utilized to update the aligned visual representation $m^v$ and aligned text representation $m^t$, enabling unimodal features to learn the correlation from each other. The process is formulated below:
\begin{equation}
    m_f^v = f_{t \rightarrow v} \times m^v,
\end{equation}
\begin{equation}
    m_f^t = f_{v \rightarrow t} \times m^t.
\end{equation}

To further improve the interaction between visual features and textual features, the cross product is done to integrate complementary semantic representations and obtain the interaction matrix $m^f$:
\begin{equation}\label{eq0}
    m^f = m_f^v \otimes m_f^t.
\end{equation}

Eventually, the multimodal interaction enhancement feature $r^f$ is obtained from the interaction matrix $m^f$ through the maxpooling operation followed by the multi-layer perceptron (MLP), which can be formulated below:

\begin{equation}
    r^f = MLP(MaxPooling(m^f)).
\end{equation}

\subsection{Adaptive Fusion-based Classification Module}
Since different features play different roles in the detection process of a specific news, it is necessary to assign specific weights to each feature for each news before performing the final fusion.

In detail, for the obtained external information enhanced text features $\tilde{r}^t$,  the multimodal interaction enhancement feature $r^f$ and the representation of the news image $r^v$, we first concatenate these three features horizontally and then compress the concatenated feature matrix using mean pooling. The attention weight of each feature is achieved by  the mean pooling result input into the fully connected layer followed by the sigmoid activation function:
\begin{equation}
    r = \tilde{r}^t \oplus r^v \oplus r^f,
\end{equation}
\begin{equation}
    \bar{r} = Mean\_Pooling(r),
\end{equation}
\begin{equation}
    a^t, a^v, a^f = sigmoid(FC(\bar{r}, W)),
\end{equation}
where $\oplus$ denotes the concatenation operation and $FC$ is the fully connected layer. At last, the final news representation $\tilde{x}$ for classification is gained from these feature weights:
\begin{equation}
    \tilde{x} = (a^t \times \tilde{r}^t) \oplus (a^v \times r^v) \oplus (a^f \times r^f),
\end{equation}
where $\oplus$ denotes the concatenation operation.

In the classification part, the final news representation $\tilde{x}$ is fed into a MLP and obtain the probability distribution $\hat{y} $ by the softmax function which is formulated as
\begin{equation}
    \hat{y} = softmax(MLP(\tilde{x})).
\end{equation}

We take the cross-entropy loss to measure the difference between the predicted probability distribution $\hat{y} $ and the true labels $y$. The loss function is shown in Equation ~\ref{eq2}:
\begin{equation}\label{eq2}
    L_d = -(y \log(\hat{y}) + (1-y) \log(1-\hat{y})).
\end{equation}

In the end, the final loss function $L$ is the combination of contrastive learning loss and the classification loss which is formulated as follows.
\begin{equation}
    L =  L_c + L_d.
\end{equation}



Finally, training with $L_c$ as the goal allows us to obtain aligned text vector $m^t$ and aligned visual vector $m^v$.

\section{Experiment}
\subsection{Experiment Setup}

\subsubsection{Datasets}

In this paper, we evaluate the proposed model ERIC-FND using the two widely used datasets Weibo and X.

Weibo dataset is constructed by Jin et al. \citeyearpar{rw-c-m-1} with the fake news in the dataset from misinformation collection by Weibo official from May 2012 to January 2016. The fake news data is from authoritative Chinese news agencies, such as Xinhua News Agency and CCTV News. The preprocessing of Weibo dataset follows Khattar's work \cite{rw-c-m-2}. News pieces with videos or without text or images are excluded. The remaining data are split into a training set and a test set in an 8:2 ratio. For the condition that one news piece related to multiple images, only the first image is selected. As shown in Table ~\ref{weibo}, the dataset consists of 9,523 news pieces, with 7,528 in the training set and 1,925 in the test set. Each news item in the dataset has its corresponding images.

X dataset is used in competition MediaEval \cite{dataset} for 
 automatically detecting fake news in various media formats on X. Similar to the Weibo dataset, only the data with both images and text are applied in the experiment. As shown in Table ~\ref{x}, the dataset has 12,514 news pieces, with 11,252 in the training set and 1,262 in the test set. 
\subsubsection{Experiment Environment}
The main parameters of the server computing environment are shown as follows:

\textbf{CPU}: AMD EPYC 7642 CPU @ 3293MHz 48C96T;

\textbf{GPU}: NVIDIA TITAN XP;

\textbf{CUDA Version}: 11.0;

\textbf{Memory}: 32GB DDR4 2666MHz ECC ×16;

\textbf{Operating System}: CentOS Linux 7 (Core)  Linux 3.10.0.

The running details are shown as follows:

\textbf{Memory needed}: 10917MiB;

\textbf{Total number of model parameters}: 132,350,153.

\begin{table}[t]
\centering
\begin{tabular}{cccc}
\toprule
         & Real News & Fake News & Total \\ \midrule
Training Set & 3783      & 3745      & 7528  \\ 
Test Set     & 999       & 996       & 1995  \\ \midrule
Total     & 4782      & 4741      & 9523  \\ 
\bottomrule
\end{tabular}
\caption{Weibo dataset distribution.}
\label{weibo}
\end{table}
\begin{table}[t]
\centering
\begin{tabular}{cccc}
\toprule
         & Real News & Fake News & Total \\ \midrule
Training Set & 5722      & 5530      & 11252  \\ 
Test Set     & 640       & 622       & 1262  \\ \midrule
Total     & 6362      & 6152      & 12514  \\ 
\bottomrule
\end{tabular}
\caption{X dataset distribution.}
\label{x}
\end{table}
\begin{table*}[t]
\centering
\begin{tabular}{ccccccccc}
\bottomrule

    \multirow{2}{*}{Dataset} & \multirow{2}{*}{Model} & \multirow{2}{*}{Accuracy}  & \multicolumn{3}{c}{Fake News} & \multicolumn{3}{c}{Real News}\\
   && & Precision & Recall & F1-score & Precision & Recall & F1-score\\
    \midrule
 \multirow{8}{*}{Weibo} & 
 att-RNN &0.772& 0.854 &0.656& 0.742& 0.720& 0.889& 0.795 \\
& EANN& 0.782 &0.827& 0.697 &0.756& 0.752 &0.863& 0.804 \\
 &MVAE& 0.824& 0.854 &0.769 &0.809& 0.802& 0.875& 0.837 \\
   & CAFE &  0.840 &0.855 &0.830& 0.842 &0.825 &0.851& 0.837 \\
   & MCAN & 0.899 &0.913& 0.889& 0.901 &0.884 &0.909&0.897\\
   & MRML & 0.897& 0.898& 0.887 &0.892 &0.896 &0.905 &0.901 \\
    &KMAGCN$_{bert}$ & 0.922 & \textbf{0.993 }& 0.851 & 0.917 & 0.869 & \textbf{0.994} & 0.927\\
    &ERIC-FND (proposed)& \textbf{0.946} & 0.985 & \textbf{0.914} &\textbf{ 0.948} &\textbf{0.908} &0.984 &\textbf{0.944} \\
\midrule
 \multirow{8}{*}{X} & 
 att-RNN & 0.664& 0.749& 0.615& 0.676& 0.589 &0.728 &0.651 \\
 &EANN &0.648& 0.810& 0.498 &0.617& 0.584& 0.759 &0.660 \\
 &MVAE &0.745& 0.801 &0.719& 0.758& 0.689& 0.777 &0.730 \\
 &CAFE &  0.806& 0.807 &0.799& 0.803& 0.805& 0.813& 0.809 \\
 &MCAN & 0.809& 0.889& 0.765& 0.822& 0.732& 0.871& 0.795\\
   & MRML&0.803& 0.821& 0.844& 0.832 &0.777& 0.747& 0.762  \\
    &KMAGCN$_{bert}$ & 0.804 & 0.787 & 0.784 & 0.785 & 0.817 & 0.819 & 0.818\\
    &ERIC-FND (proposed) & \textbf{0.945} & \textbf{0.987 }&\textbf{ 0.910} &\textbf{ 0.947} & \textbf{0.905} & \textbf{0.986} & \textbf{0.944} \\
\bottomrule
\end{tabular}
\caption{Comparative experiment results on Weibo dataset and X dataset.}
\label{compare}
\end{table*}

\begin{table*}[t]
\centering

\begin{tabular}{ccccc}
\bottomrule
    Model & Accuracy  & Precision & Recall & F1-score \\
    \midrule

  ERIC-FND   w/o A & 0.938 & 0.939 & 0.938 & 0.938\\
   ERIC-FND w/o M & 0.925 & 0.925 & 0.932 & 0.925 \\
   ERIC-FND  w/o E & 0.926 & 0.933 & 0.926 & 0.926 \\
    ERIC-FND & \textbf{0.946} & \textbf{0.949} & \textbf{0.946} & \textbf{0.946} \\

\bottomrule
\end{tabular}
\caption{Ablation test results on Weibo dataset. A represents adaptive fusion-based classification module. M represents multimodal information interactive enhancement module. E represents external information enhancement module. }
\label{ab}
\end{table*}

\subsubsection{Parameter Settings and Implementation Details}
The settings of used parameters and the implementation 
 details are described in the Appendix Section. The random seed is set and each experiment is run five times to ensure the reproduciblity. 
 
\subsubsection{Baselines}

We select several models in recent multimodal fake news detection studies as baseline models and the variants of proposed model ERIC-FND for comparison.

Recent multimodal fake news detection studies include
\textbf{att-RNN} \cite{attrnn}, \textbf{EANN } \cite{eann}, \textbf{MVAE} \cite{mvae}, \textbf{CAFE} \cite{cafe}, \textbf{MCAN} \cite{wu-etal-2021-multimodal},  \textbf{MRML} \cite{10096188} and  \textbf{KMAGCN}\cite{10.1145/3451215}.

\begin{figure}[t]
\centering
\includegraphics[width=0.45\textwidth]{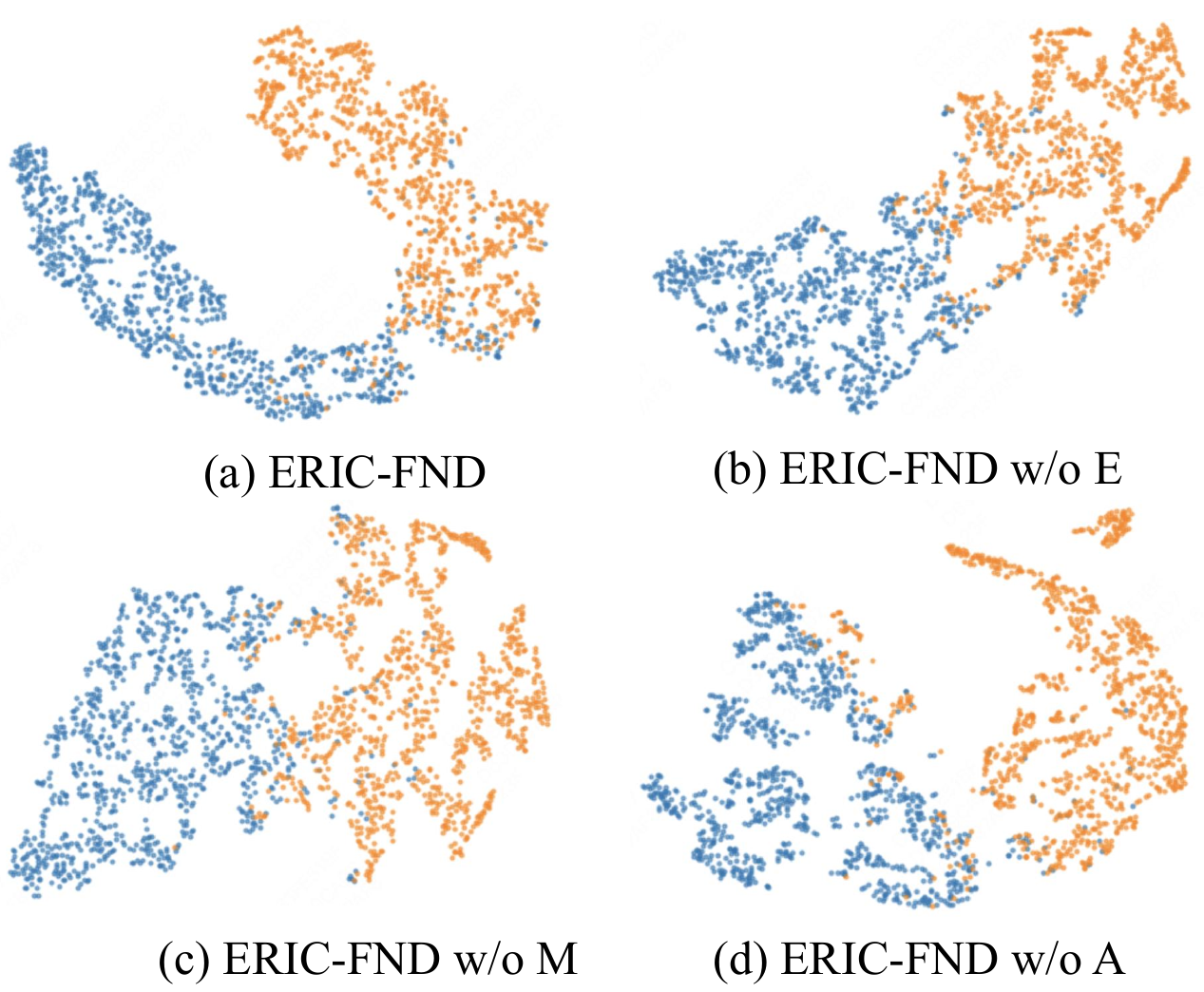} 
\caption{The t-SNE feature visualization of proposed model ERIC-FND and its variants on Weibo dataset.   }
\label{visual}
\end{figure}

Variants of proposed model ERIC-FND:

\textbf{ERIC-FND w/o A}: The variant of proposed model ERIC-FND in this paper without adaptive fusion-based classification module. The final news representation is the simple concatenation of external information enhanced text features, the multimodal interaction enhancement feature and the representation of the news image. 

\textbf{ERIC-FND w/o M}: The variant of proposed model ERIC-FND in this paper without multimodal information interactive enhancement module.

\textbf{ERIC-FND w/o E}: The variant of proposed model ERIC-FND in this paper without external information enhancement module.

All models are evaluated with the widely used metrics: accuracy, precision,
recall and F1 score.

\subsection{Comparative Experiment}
The comparative experiment results are presented in Table ~\ref{compare} and the best results are in bold. Our proposed model ERIC-FND mostly outperforms the SOTA multimodal fake news detection models in accuracy, precision, recall and F1-score with different categories. Therefore, the comparative experiment results demonstrate that our proposed model ERIC-FND effectively exploits multimodal news information and the external information enhancement helps deepen the model understanding of textual semantics with obtained external reliable information. What is more, the adaptive fusion makes contributions to the selection of key features for detection.

\subsection{Ablation Study}

The results of ablation test are shown in Table ~\ref{ab}. It can be seen from the table that the proposed model outperforms all the variants on both datasets. The removal of different module has various impact on the model performance. To be specific, after removing the adaptive fusion-based classification module, the model performance drops 0.8\% accuracy on both datasets, which indicates the adaptive fusion is able to help extract key features among all the obtained feature representation. With the multimodal information interactive enhancement module, the model separately achieves an improvement of 2.1\% and 1.6\% in Weibo and X dataset compared to the variant model, which proves that the multimodal information interactive enhancement enables to capture the deep semantic interaction of multimodal information. As to the addition of external information enhancement module, there are increases of  2\% and 3.1\% in accuracy on Weibo and X datasets which shows that extracted entity description helps enhance the news textual representation. 

\subsection{Visualization Analysis}

The visualization analysis of the obtained news representations before being input into the classifier is conducted to further validate the effectiveness of our proposed model. Specifically, we apply the t-distributed Stochastic Neighbor Embedding (t-SNE) method \cite{JMLR:v9:vandermaaten08a} on the Weibo dataset to perform dimensionality reduction on the obtained news representations before being input into the classifier. The dimensionality-reduced data are then visualized which provides a clear observation of the distribution of news data points with different labels in low dimensionality space. Figure ~\ref{visual} shows the visualization results of proposed model ERIC-FND and its variant models on  Weibo dataset. In the figure, blue points represent real news, and orange points represent fake news.

As shown in the figure, the news feature representations obtained by ERIC-FND are more distinct with different labels in the low dimensionality, which indicates that the proposed model has the capability of effectively differentiating different categories of news pieces. This result demonstrates that the proposed model is able to largely enhance the final news representation features, further proving the model effectiveness.

\section{Conclusion}
In this paper, we propose the ERIC-FND model, which employs entity-based external reliable information to enhance the understanding of textual content and multimodal information interactive semantic enhancement to improve cross-modality learning. Finally, an adaptive fusion method is utilized to optimize the contribution of each news feature to the final classification. Experimental results demonstrate that ERIC-FND outperforms SOTA models and its variants on two datasets, indicating the effectiveness of the ERIC-FND architecture.

\section{Acknowledgements}
This work is supported by National Natural Science Foundation of China under Grants No.62472092, No.62172089, No.62172090, No.62106045. Natural Science Foundation of Jiangsu province under Grants No.BK20241751. Jiangsu Provincial Key Laboratory of Computer Networking Technology. Jiangsu Provincial Key Laboratory of Network and Information Security under Grants No.BM2003201, and Key Laboratory of Computer Network and Information Integration of Ministry of Education of China under Grants No.93K-9, Nanjing Purple Mountain Laboratories. Start-up Research Fund of Southeast University under Grants No.RF1028623097. We thank the Big Data Computing Center of Southeast University for providing the facility support on the numerical calculations.

\end{document}